\documentclass[10pt,conference]{IEEEtran}

\usepackage{cite}
\usepackage{amsmath,amssymb,amsfonts}
\usepackage{algorithmic}
\usepackage{graphicx}
\usepackage{textcomp}
\usepackage{comment}
\usepackage{url}
\usepackage{hyperref}
\usepackage{diagbox}
\usepackage{siunitx}

\begin{document}

\title{Refined CMA-ES for CEC 2026 Bound-Constrained Single-Objective Optimization}

\author{
    \IEEEauthorblockN{
        Adam Stelmaszczyk\IEEEauthorrefmark{1},
        Rafał Biedrzycki\IEEEauthorrefmark{1},
        Jarosław Arabas\IEEEauthorrefmark{1}
    }
    \IEEEauthorblockA{\IEEEauthorrefmark{1}Institute of Computer Science, Warsaw University of Technology, Warsaw, Poland\\
    \{adam.stelmaszczyk, rafal.biedrzycki, jaroslaw.arabas\}@pw.edu.pl}
}

\maketitle

\let\thefootnote\relax\footnotetext{
    \footnotesize © 2026 IEEE. Personal use of this material is permitted. Permission from IEEE must be obtained for all other uses, in any current or future media, including reprinting/republishing this material for advertising or promotional purposes, creating new collective works, for resale or redistribution to servers or lists, or reuse of any copyrighted component of this work in other works.\\
    This is the author's accepted manuscript. The final published version is available at IEEE Xplore: \url{https://doi.org/10.1109/CEC69861.2026.11727142}.
}

\begin{abstract}
We present a simplified and parameter-adjusted variant of CMA-ES designed for the CEC 2026 Bound-Constrained Single-Objective Optimization competition. Unlike typical algorithmic developments that increase complexity by adding mechanisms, Refined CMA-ES systematically removes components that do not improve performance, such as the flat‑land escape, while retaining standard features that provide measurable benefits. Adjusting key parameters---population size and initial step-size---yields substantial gains, outperforming the baseline CMA-ES and several state-of-the-art competitors. The paper highlights the practical relevance of individual CMA-ES components under the CEC 2026 ranking, which equally weights solution accuracy and convergence speed, demonstrating that simplification combined with improved parameter settings can lead to superior performance.
\end{abstract}


\section{Introduction}

In evolutionary computation, algorithmic development typically follows an additive paradigm: a baseline method is taken and additional mechanisms are progressively introduced in order to improve performance. Examples include restart strategies, adaptive population sizes or hybridization. Over time, this process often leads to increasingly complex algorithms, with the effects of individual components not always clearly assessed.

In this work, we adopt the opposite approach. Instead of adding new components to CMA-ES, we systematically analyze the impact of its existing mechanisms using the CEC 2026 benchmark. The goal is not to increase algorithmic complexity, but to identify which components are actually beneficial in this specific evaluation setting, and which can be removed without harming performance---or even improving it.

From this perspective, the CEC 2026 benchmark is treated not only as a competition platform, but also as a diagnostic tool for performing an ablation study of CMA-ES. By measuring the effect of removing selected components and adjusting key parameters, we obtain insights into the practical relevance of several design choices in the original algorithm.

An important aspect of this study is the specific nature of the CEC 2026 ranking itself. Unlike most traditional benchmarks that focus solely on final solution quality, the CEC 2026 ranking assigns equal importance to two criteria: solution quality and convergence speed. Half of the total score is determined by the final error values, while the other half measures how quickly an algorithm reaches the solution it ultimately reports as the result of the optimization process, according to the official evaluation protocol \cite{cec26}.

This hybrid evaluation scheme significantly alters the optimization landscape. Mechanisms that are beneficial for long-term fine-tuning of solutions may become less relevant, or even detrimental, when early convergence is heavily rewarded. As a result, some classical components of CMA-ES, originally designed for robust performance, may no longer be useful under this ranking protocol. This makes the CEC 2026 benchmark particularly well suited for analyzing which parts of CMA-ES are truly effective in practice under time-sensitive evaluation.

The main contributions of this paper are:
\begin{itemize}
    \item an assessment of the covariance matrix decomposition delay in CMA-ES, highlighting its limited application,
    \item an empirical evaluation of the flat‑land escape mechanism, which was found to degrade performance on CEC 2026 benchmark,
    \item an empirical evaluation indicating that the $h_\sigma$ correction, which stalls evolution path adaptation when the step-size becomes too small, provides a small performance improvement on the CEC 2026 benchmark,
    \item a demonstration that parameter settings alone yield substantial performance gains,
    \item the proposal of Refined CMA-ES (R-CMA-ES), a simplified and parameter-adjusted variant that outperforms the baseline CMA-ES and several state-of-the-art competitors.
\end{itemize}

The remainder of this paper is organized as follows. Section II describes our baseline, the CMA-ES algorithm. Section III details the methodology used to measure the impact of each modification. Section IV introduces R-CMA-ES, provides a detailed description of the modifications, and defines the algorithm variations. Section V presents pairwise and total-point comparisons of these variants. Section VI compares R-CMA-ES with selected existing methods. Finally, Section VII discusses implementation details and reports on the algorithmic complexity of R-CMA-ES.

\section{CMA-ES}

As a base algorithm, we use the state-of-the-art Covariance Matrix Adaptation Evolution Strategy (CMA-ES) as described in \cite{Hansen2016Tutorial}. CMA-ES is a population-based, derivative-free optimization algorithm designed for continuous black-box problems. At each generation, a population of candidate solutions is sampled from a multivariate normal distribution
\[
x_i \sim \mathcal{N}(m, \sigma^2 C),
\]
where $m$ is the current mean, $\sigma$ is the global step-size, and $C$ is the covariance matrix.

The mean vector $m$ is updated using a weighted recombination of the best individuals in the population, and the covariance matrix $C$ is adapted over time, enabling an implicit form of second-order optimization.

In addition to mean and covariance matrix adaptation, CMA-ES employs step-size control based on the evolution path mechanism, which adjusts $\sigma$ to regulate the overall scale of the search steps. Beyond these core principles, several auxiliary mechanisms have been incorporated into the standard algorithm over time, including covariance matrix decomposition delay, flat-land escape, and the $h_\sigma$ correction.

\section{Methodology}

The CEC 2026 ranking \cite{cec26} evaluates algorithms based on both solution accuracy and convergence speed, assigning equal weight to each criterion. For pairwise comparisons, each algorithm is run for 25 independent trials on each of the 29 benchmark functions. This yields $50 \cdot 49 / 2 = 1225$ pairwise trial comparisons per function, resulting in a maximum of 35525 points per objective. Combining accuracy and speed objectives, a total of 71050 points can be awarded for each pair of algorithms.

Table \ref{tab:pairwise} reports pairwise differences, computed as the difference between the CEC 2026 ranking points achieved by algorithm~1 and algorithm~2. Positive values indicate that the first algorithm outperformed the second. In contrast, Tables \ref{tab:direct} and \ref{tab:all} present the CEC 2026 ranking for several algorithms, where total points across all functions and objectives are aggregated to rank all compared algorithms simultaneously.

This methodology allows us to quantify both absolute performance and relative differences between algorithms, providing a clear view of which modifications and parameter settings yield measurable improvements under the hybrid accuracy-speed evaluation.

\section{Refined CMA-ES}

R-CMA-ES is a parameter-adjusted CMA-ES in which the covariance matrix decomposition delay and flat-land escape mechanisms have been removed. The following subsections provide a detailed description of these modifications and define the A, B, C algorithm variations.

\subsection{No covariance matrix decomposition delay}

In CMA-ES, the covariance matrix decomposition is delayed by $\max(1, \lfloor 0.1 \cdot n \cdot c_{cov} \rfloor)$ generations, where the learning coefficient $c_{cov}$ depends on the problem dimension $n$. With the default CMA-ES parameters, $\lambda = 4 + \lfloor 3 \log(n) \rfloor$, this delay is greater than one generation only for dimensions $n \ge 190$. When $\lambda \ge n$, decomposition occurs every generation, resulting in a per-generation computational complexity of $O(n^3)$. Therefore, for $n < 190$, the delay mechanism is effectively inactive, and we simplify the implementation by removing it without affecting the observed behavior.

\subsection{No flat-land escape}

Step-size $\sigma$ is increased when multiple individuals share the same fitness, aiming to escape flat regions of the fitness landscape more quickly. However, as noted in \cite{Hansen2016Tutorial}, this mechanism has certain drawbacks: “This method interferes with the termination criterion \texttt{TolFun}. In practice, observation of a flat fitness should rather be a termination criterion and consequently lead to a reconsideration of the objective function formulation.” By removing the flat-land escape, these issues are avoided, and we observed an improvement in overall performance on the CEC 2026 benchmark.

\subsection{Non-default parameter settings}

By default, CMA-ES uses a population size $\lambda = 4 + \lfloor 3 \log(n) \rfloor$ and an initial step-size $\sigma = 0.5$. Inspired by RB-IPOP-CMA-ES \cite{cec17rb}, we adjusted these two parameters. First, the population size was increased to $\lambda = 4n$, corresponding to an increase from 14 to 120 in the 30-dimensional CEC 2026 benchmark. Second, the initial step-size $\sigma$ was increased from 0.5 to 7. Both changes result in substantial performance gains, considerably larger than removing the flat-land escape, as shown later.

\subsection{Other modifications}

In addition to the modifications reported above, we also examined the impact of removing the correction of the evolution path adaptation $h_\sigma$, which is a standard component of the original CMA-ES. In contrast to decomposition delay and flat-land escape, removing $h_\sigma$ caused a slight performance decrease. Consequently, $h_\sigma$ was retained in the final R-CMA-ES.

Furthermore, we evaluated several mechanisms commonly reported in the CMA-ES literature, such as IPOP restarts \cite{IPOP}, bound constraint handling by resampling \cite{CMAES_BCHMs}, hybridization with a local optimizer (L-BFGS-B), the use of midpoint evaluation, and starting point presampling \cite{cec17rb}. None of these extensions led to performance improvements under the CEC 2026 evaluation. In particular, restart strategies resulted in a minor improvement in accuracy, but this was offset by a substantial drop in convergence speed, which is heavily weighted in the CEC 2026 ranking.\\

Table \ref{tab:pairwise} presents pairwise comparisons of the CMA-ES variants using the CEC 2026 ranking. Algorithm A is omitted, as its observed behavior is identical to the baseline CMA-ES in 30 dimensions.

\begin{table}[!ht]
\centering
\renewcommand{\arraystretch}{1.2}
\caption{Pairwise comparisons of CMA-ES variants using CEC 2026 ranking points}
\begin{tabular}{|c|S[table-format=6.1, group-minimum-digits=4]|
                  S[table-format=6.1, group-minimum-digits=4]|
                  S[table-format=6.1, group-minimum-digits=4]|}
\hline
\diagbox{algorithm 1}{algorithm 2} & {CMA-ES} & {AB} & {AC} \\ \hline
AB & 1615 & {---} & {---} \\ \hline
AC & 27816 & 27529 & {---} \\ \hline
\textbf{R-CMA-ES (ABC)} & 28308 & 28069 & 1275 \\ \hline
\end{tabular}
\label{tab:pairwise}
\end{table}

Variant AB (no delay + no escape) won against the original CMA-ES by 1615 ranking points. AC (no delay + parameter settings) outperformed the original CMA-ES by a much larger margin of 27816 points, showing that parameter settings alone provide a significant performance improvement. This is further confirmed by comparing AC and AB: AC won by 27529 points. Finally, R-CMA-ES (no delay + no escape + parameter settings) beats AC by 1275 points, a much smaller margin.

Table \ref{tab:direct} presents the CEC 2026 ranking for the considered CMA-ES variants based on total CEC 2026 points. R-CMA-ES takes first place with 100601 points. The AC variant follows in second place with 98810 points, trailing by 1791 points, confirming that the improved parameter choices account for most of the performance gain. The AB variant ranks third with 44783 points, showing that the removal of the flat-land escape provides only a small gain of 1878 points compared to the baseline CMA-ES.

\begin{table}[!ht]
\centering
\renewcommand{\arraystretch}{1.2}
\caption{CEC 2026 ranking of considered CMA-ES variants}
\begin{tabular}{|c|c|S[table-format=6.1]|}
\hline
place & algorithm & {points} \\ \hline
1 & \textbf{R-CMA-ES (ABC)} & 100601 \\ \hline
2 & AC & 98810 \\ \hline
3 & AB & 44783 \\ \hline
4 & CMA-ES & 42905 \\ \hline
\end{tabular}
\label{tab:direct}
\end{table}

Table \ref{tab:wilcoxon} presents the mean of the minimum error values (Min\_EV) obtained across 25 independent trials for 30 dimensional problems. The "Test" column reports the result of the Wilcoxon signed-rank test comparing the 25 Min\_EV values. "=" indicates no statistically significant difference (p~=~0.05). "+" indicates that R-CMA-ES performed significantly better, whereas "--" indicates it performed worse. Overall, in 30 dimensions, R-CMA-ES performed significantly better on 23 out of 29 functions and worse on one function (F3).

\begin{table}[!ht]
\centering
\renewcommand{\arraystretch}{1.2}
\caption{Minimum error values averaged over 25 trials in 30 dimensions}
\begin{tabular}{|c|c|c|c|}
\hline
Function & R-CMA-ES & CMA-ES & Test \\
\hline
1 & \textbf{0.0e+00} & 1.1e-14 & + \\
2 & 1.0e+02 & 1.0e+02 & = \\
3 & 1.6e+02 & \textbf{1.4e+02} & -- \\
4 & \textbf{1.1e+02} & 7.8e+02 & + \\
5 & \textbf{1.0e+02} & 1.9e+02 & + \\
6 & \textbf{1.4e+02} & 2.8e+03 & + \\
7 & \textbf{1.1e+02} & 7.0e+02 & + \\
8 & \textbf{1.0e+02} & 1.5e+04 & + \\
9 & \textbf{2.1e+03} & 5.3e+03 & + \\
10 & \textbf{2.0e+02} & 3.1e+02 & + \\
11 & 1.5e+03 & 1.6e+03 & = \\
12 & \textbf{1.9e+02} & 2.4e+03 & + \\
13 & \textbf{2.1e+02} & 3.0e+02 & + \\
14 & \textbf{3.6e+02} & 4.7e+02 & + \\
15 & \textbf{3.1e+02} & 9.3e+02 & + \\
16 & \textbf{1.9e+02} & 6.3e+02 & + \\
17 & 3.3e+02 & 3.3e+02 & = \\
18 & \textbf{2.0e+02} & 3.2e+02 & + \\
19 & \textbf{4.2e+02} & 1.4e+03 & + \\
20 & \textbf{3.1e+02} & 6.2e+02 & + \\
21 & \textbf{1.1e+03} & 6.1e+03 & + \\
22 & \textbf{4.6e+02} & 2.1e+03 & + \\
23 & \textbf{5.2e+02} & 6.2e+02 & + \\
24 & \textbf{4.9e+02} & 4.9e+02 & + \\
25 & 9.8e+02 & 1.3e+03 & = \\
26 & \textbf{6.1e+02} & 9.0e+02 & + \\
27 & 4.5e+02 & 4.4e+02 & = \\
28 & \textbf{5.6e+02} & 1.1e+03 & + \\
29 & \textbf{2.3e+03} & 2.7e+03 & + \\
\hline
Wins & 23 & 1 \\
\cline{1-3}
\end{tabular}
\label{tab:wilcoxon}
\end{table}

\section{Comparison with selected existing methods}

Table \ref{tab:all} presents a comparison of Refined CMA-ES with the baseline CMA-ES and three other notable algorithms: L-SRTDE \cite{L-SRTDE}, EA4EigSimpTowardsIDE \cite{EA4EigSimpTowardsIDE}, and RB-IPOP-CMA-ES \cite{cec17rb}. L-SRTDE was the winner of the CEC 2024 competition, EA4EigSimpTowardsIDE is an improved simplification of EA4Eig, which won CEC 2022, and RB-IPOP-CMA-ES reached 7th place in CEC 2017.  

\begin{table}[!ht]
\centering
\renewcommand{\arraystretch}{1.2}
\caption{Comparison with selected existing methods on the CEC 2026 benchmark}
\begin{tabular}{|c|c|S[table-format=6.1]|}
\hline
place & algorithm & points \\ \hline
1 & \textbf{R-CMA-ES} & 109816 \\ \hline
2 & L-SRTDE & 98387 \\ \hline
3 & RB-IPOP-CMA-ES & 85123 \\ \hline
4 & CMA-ES & 79213 \\ \hline
5 & EA4EigSimpTowardsIDE & 76960 \\
\hline
\end{tabular}
\label{tab:all}
\end{table}

Refined CMA-ES achieved the highest score, outperforming the runner-up L-SRTDE by 11428 points. The relatively lower performance of RB-IPOP-CMA-ES can be attributed to its use of IPOP restarts: restart strategies improve the accuracy but cause a substantial drop in convergence speed, which is heavily weighted in the CEC 2026 ranking. Overall, these results demonstrate that selective simplification combined with improved initial parameter settings yields superior performance compared to both the baseline CMA-ES and other state-of-the-art methods.

\section{Implementation notes and algorithm complexity}

The full source code is available at \href{https://github.com/AdamStelmaszczyk/refined-cma-es}{https://github.com/AdamStelmaszczyk/refined-cma-es}. The experiments were conducted on a Linux Debian system equipped with an Intel Xeon CPU E7-4830 v3, with a base clock of 2.1 GHz and turbo frequency up to 2.7 GHz. R-CMA-ES is a single-threaded R implementation based on the \texttt{cmaes} package \cite{cmaesRpackage}. The CEC 2026 benchmark relies on CEC 2017 functions, for which we used the \texttt{cec2017} package \cite{CEC2017Rpackage}. The algorithmic complexity was assessed following the methodology described in \cite{cec26}. Table \ref{tab:complexity} reports the timing results. $T_1$ denotes the time required for 10000 function evaluations, averaged over the 29 CEC functions. $T_2$ denotes the time required to run R-CMA-ES with a budget of 10000 evaluations, also averaged over the 29 functions. In 30 dimensions, approximately 90\% of the total runtime is spent on function evaluations.

\begin{table}[!ht]
\centering
\renewcommand{\arraystretch}{1.2}
\caption{Algorithm complexity}
\begin{tabular}{|c|c|c|}
\hline
$T_1$ & $T_2$ & $(T_2 - T_1)/T_1$ \\ \hline
3.32s & 3.66s & 0.1 \\ \hline
\end{tabular}
\label{tab:complexity}
\end{table}

\section{Conclusions}

This paper introduced Refined CMA-ES, a simplified and parameter-adjusted variant of CMA-ES, which achieves superior performance on the CEC 2026 benchmark compared to the state-of-the-art methods. Refined CMA-ES outperforms L-SRTDE, the winner of CEC 2024, demonstrating that simplification combined with improved parameter settings can yield substantial gains. The study also shows that certain CMA-ES mechanisms, such as covariance matrix decomposition delay and flat‑land escape, are not beneficial under this evaluation, while adjusting key parameters---population size and initial step-size---has a strong positive impact on both accuracy and convergence speed.

\section*{Acknowledgment}

We gratefully acknowledge Polish high-performance computing infrastructure PLGrid (HPC Center: CI TASK) for providing computer facilities and support within computational grant no. PLG/2025/018266.

\bibliographystyle{IEEEtran}
\bibliography{references}

\end{document}